\documentclass[letterpaper]{article} 
\usepackage{aaai23}  
\usepackage{times}  
\usepackage{helvet}  
\usepackage{courier}  
\usepackage[hyphens]{url}  
\usepackage{graphicx} 
\usepackage{natbib}  
\usepackage{caption} 
\usepackage{algorithm}
\usepackage{algorithmic}

\usepackage{amsmath,amssymb}
\usepackage{subcaption}
\usepackage{bbding}
\usepackage{makecell}
\usepackage{multirow}
\usepackage{booktabs}
\usepackage{colortbl}
\usepackage[switch]{lineno}
\usepackage{dsfont}

\definecolor{lightgray}{rgb}{0.86, 0.86, 0.86}
\usepackage{newfloat}
\usepackage{listings}
\DeclareCaptionStyle{ruled}{labelfont=normalfont,labelsep=colon,strut=off} 
\floatstyle{ruled}
\newfloat{listing}{tb}{lst}{}
\floatname{listing}{Listing}
\nocopyright

\title{End-to-End Zero-Shot HOI Detection via \\Vision and Language Knowledge Distillation}
\author{
    Mingrui Wu\textsuperscript{\rm 1,\rm 2${\dag}$}\thanks{ Intern at Youtu Lab, Tencent. ${\dag}$ Equal Contribution. Works completed at Tencent. ${\ddag}$ Corresponding Author.}, 
    	Jiaxin Gu\textsuperscript{\rm 3${\dag}$},
    	Yunhang Shen\textsuperscript{\rm 2}, 
    	Mingbao Lin\textsuperscript{\rm 2},\\ 
    	Chao Chen\textsuperscript{\rm 2}, 
    	Xiaoshuai Sun\textsuperscript{\rm 1,\rm 4,\rm 5${\ddag}$}\\
}
\affiliations{
        \textsuperscript{\rm 1}MAC Lab, School of Informatics, Xiamen University.
    	\textsuperscript{\rm 2}Youtu Lab, Tencent. \\
    	\textsuperscript{\rm 3}VIS, Baidu Inc.
    	\textsuperscript{\rm 4}Institute of Artificial Intelligence, Xiamen University. \\
    	\textsuperscript{\rm 5}Fujian Engineering Research Center of Trusted Artificial Intelligence Analysis and Application, Xiamen University.\\
    	{\tt\small  mingrui0001@gmail.com, gujiaxin02@baidu.com, shenyunhang01@gmail.com, 
    	linmb001@outlook.com\\
    	\tt\small  aaronccchen@tencent.com, xssun@xmu.edu.cn
    	}

}

\begin{document}

\maketitle

\begin{abstract}
Most existing Human-Object Interaction~(HOI) Detection methods rely heavily on full annotations with predefined HOI categories, which is limited in diversity and costly to scale further.
We aim at advancing zero-shot HOI detection to detect both seen and unseen HOIs simultaneously.
The fundamental challenges are to discover potential human-object pairs and identify novel HOI categories.
To overcome the above challenges, we propose a novel \textbf{E}nd-to-end zer\textbf{o}-shot HO\textbf{I} \textbf{D}etection (\textbf{EoID}) framework via vision-language knowledge distillation.
We first design an \emph{Interactive Score} module combined with a \emph{Two-stage Bipartite Matching} algorithm to achieve interaction distinguishment for human-object pairs in an action-agnostic manner.
%
Then we transfer the distribution of action probability from the pretrained vision-language teacher as well as the seen ground truth to the HOI model to attain zero-shot HOI classification.
Extensive experiments on HICO-Det dataset demonstrate that our model discovers potential interactive pairs and enables the recognition of unseen HOIs.
%
Finally, our EoID outperforms the previous SOTAs under various zero-shot settings.
Moreover, our method is generalizable to large-scale object detection data to further scale up the action sets.
The source code is available at: \url{https://github.com/mrwu-mac/EoID}.
\end{abstract}

\section{Introduction}

\begin{figure}[t]
\centering
\includegraphics[width=0.98\columnwidth]{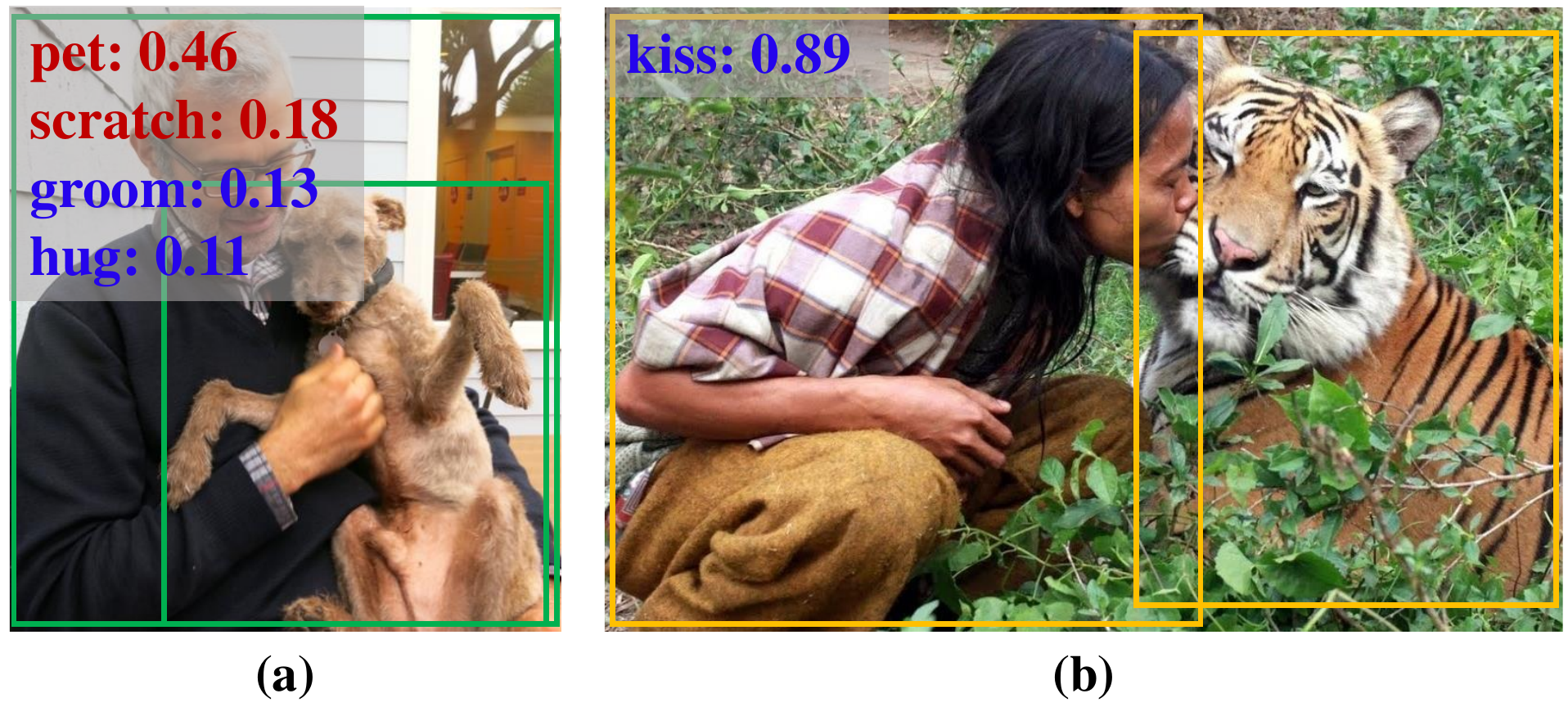} 
\caption{CLIP shows promising zero-shot ability in detecting \textit{unseen} actions ({\color{red}red}) and \textit{unseen} objects (\emph{e.g.} \textit{tiger}) with \textit{seen} actions ({\color{blue}blue}). The \textit{unseen} and \textit{seen} here denote the divisions under the zero-shot setting.}
\vspace{-0.6cm}
\label{fig:intro}
\end{figure}

The task of Human-Object Interaction~(HOI) detection aims to detect $\langle human,$ $verb, object \rangle$ triplets, which simultaneously localizes human-object pairs and identifies the corresponding interactive actions.
HOI detection plays an important role in many downstream visual understanding tasks, especially for human-centric scenes, such as Image Captioning~\cite{li2017scene} and Visual Question Answering~\cite{goyal2017making}.

Most of the existing works simply focus on improving action classification performances for predefined HOI categories.
However, they suffer from two major weaknesses: 1) excessive cost for new HOI dataset construction and 2) a lack of generalization for unseen actions.
%
%
Previous works~\cite{shen2018scaling, bansal2020detecting, liu2020consnet, hou2021detecting} attempted to overcome the above drawbacks via zero-shot learning.
Most of them are devoted to improving human-object visual representation~\cite{hou2021detecting} and introducing language model~\cite{bansal2020detecting, liu2020consnet}, ignoring the implicit relations between vision and language.

As a recent technological breakthrough, CLIP~\cite{radford2021learning} performs contrastive learning on $400$ million image-text pairs collected from the Web and shows impressive zero-shot transferability on over $30$ classification datasets. Some recent works also successfully transfer the pretrained CLIP model to various downstream tasks such as object detection~\cite{gu2021open}, text-driven image manipulation~\cite{patashnik2021styleclip} and semantic segmentation~\cite{zhou2021denseclip}. Compared to text embeddings simply extracted from pure language models, the text embeddings learned jointly with visual images can better encode the visual similarity between concepts~\cite{gu2021open}.
%
Inspired by this, we attempt to transfer vision and language (V$\&$L) knowledge of CLIP into the zero-shot HOI task.
%
As shown in Fig.\,\ref{fig:intro}, CLIP has the ability to identify certain unseen actions (Fig.\,\ref{fig:intro}(a)), and also discover unseen objects with seen action interaction (Fig.\,\ref{fig:intro}(b)).
%
%
%
Recent GEN-VLKT~\cite{liao2022gen} applies CLIP to HOI detection task. However, GEN-VLKT is limited in known human-object pairs and fails to deal with the potential interactive pairs. 
In addition, its global image-level distillation inevitably introduces noise when there exist multiple interactions in one image. It requires local region-level distillation for accuracy and robustness.

%
To this end, we propose EoID, an end-to-end zero-shot HOI detection framework to detect unseen HOI pairs by distilling the knowledge from CLIP.
We first design a novel \emph{Interactive Score} (\textbf{IS}) module with a \emph{Two-stage Bipartite Matching} algorithm to discover potential action-agnostic interactive human-object pairs. Then we distill interactive knowledge from CLIP to teach the HOI model to identify unseen actions. Specifically, the probability distribution of the actions is obtained by CLIP given the cropped union regions of each human-object pairs, with predefined HOI prompts. Finally, the HOI model learns from the distilled probability distribution as well as the ground truth actions.
%
%
We evaluate the proposed method on HICO-Det~\cite{chao2018learning} benchmark under unseen action (UA) and unseen action-object combination (UC) settings. Extensive experiments validate that our framework can detect potential interactive human-object pairs. 
And the results show that our approach outperforms previous SOTAs under various zero-shot settings.
In addition, our method can generalize to object detection datasets and obtain $47.15\%$ $m$AP on unseen actions of V-COCO~\cite{gupta2015visual} only with the bounding boxes from MS-COCO~\cite{lin2014microsoft}.

To summarize, our contributions are:
\begin{itemize}
\item We propose an end-to-end zero-shot HOI detection framework which attains zero-shot HOI classification via V\&L knowledge distillation.
\item We succeed in detecting potential action-agnostic interactive human-object pairs by applying an \emph{Interactive Score} module combined with a \emph{Two-stage Bipartite Matching} algorithm, the effectiveness of which has been validated through extensive experiments.
\item Experiments show that EoID is capable of detecting HOIs with unseen HOI categories and outperforms previous SOTA under zero-shot settings. Moreover, our method is able to generalize to object detection datasets only with bounding boxes, which further scales up the action sets.
\end{itemize}

\section{Related Works}
\subsection{Human-Object Interaction Detection}
Most previous works on HOI detection can be categorized into two groups: two-stage~\cite{chao2018learning, gao2018ican, li2019transferable, kim2020detecting} and one-stage~\cite{gkioxari2018detecting, liao2020ppdm, wang2020learning, kim2020uniondet, kim2021hotr, zou2021end, tamura2021qpic, zhang2021mining} methods.
%
%
%
Following this paradigm, we build our framework on the transformer-based approach to achieve zero-shot HOI detection. We also replace the one-stage Hungarian matching algorithm~\cite{kuhn1955hungarian, carion2020end, tamura2021qpic, zhang2021mining} with a novel two-stage matching algorithm for detecting 
potential interactive human-object pairs.

\subsection{Knowledge Distillation from CLIP}
CLIP~\cite{radford2021learning} adopts contrastive learning to jointly train image-text embedding models on large-scale image-text pairs collected from the internet and has shown promising zero-shot transferability. It inspires subsequent studies to transfer the vision and language knowledge to various downstream tasks such as object detection~\cite{gu2021open}, text-driven image manipulation~\cite{patashnik2021styleclip}, video clip retrieval~\cite{luo2021clip4clip} and semantic segmentation~\cite{zhou2021denseclip, rao2021denseclip}.
%
Recent GEN-VLKT~\cite{liao2022gen} for the first time applies CLIP to HOI detection task, which transfers the knowledge of CLIP by image-level feature distillation. However, it fails to deal with multiple human-object interactions in one image. Different from previous attempts, we propose to use region-level distillation by distilling soft action probability from CLIP.

\subsection{Zero-Shot Learning on HOI detection}

Zero-shot learning aims at classifying categories that are not seen during training.
Previous works implement zero-shot learning on HOI detection task from three scenarios: unseen combination scenario~(UC)~\cite{shen2018scaling, bansal2020detecting, hou2021detecting, liu2020consnet}, unseen object scenario~(UO)~\cite{bansal2020detecting, hou2021detecting, liu2020consnet} and unseen action scenario~(UA)~\cite{liu2020consnet}.
ConsNet~\cite{liu2020consnet} performs zero-shot HOI detection for the three scenarios by learning from a consistency graph along with word embeddings. 
\cite{wang2022learning} develop a transferable HOI detector via joint visual-and-text modeling. RLIP~\cite{yuan2022rlip} propose relational Language- Image pre-training to improve zero-shot, few-shot and fine-tuning HOI detection performance. However, they show limited zero-shot capability due to limited HOI datasets.
%
GEN-VLKT~\cite{liao2022gen} distills CLIP knowledge into known human-object pairs to attain zero-shot learning. In contrast, our proposed method is able to attain zero-shot learning on known human-object pairs as well as potential interactive pairs.

\begin{figure*}[!t]
\centering
\includegraphics[width=0.85\textwidth]{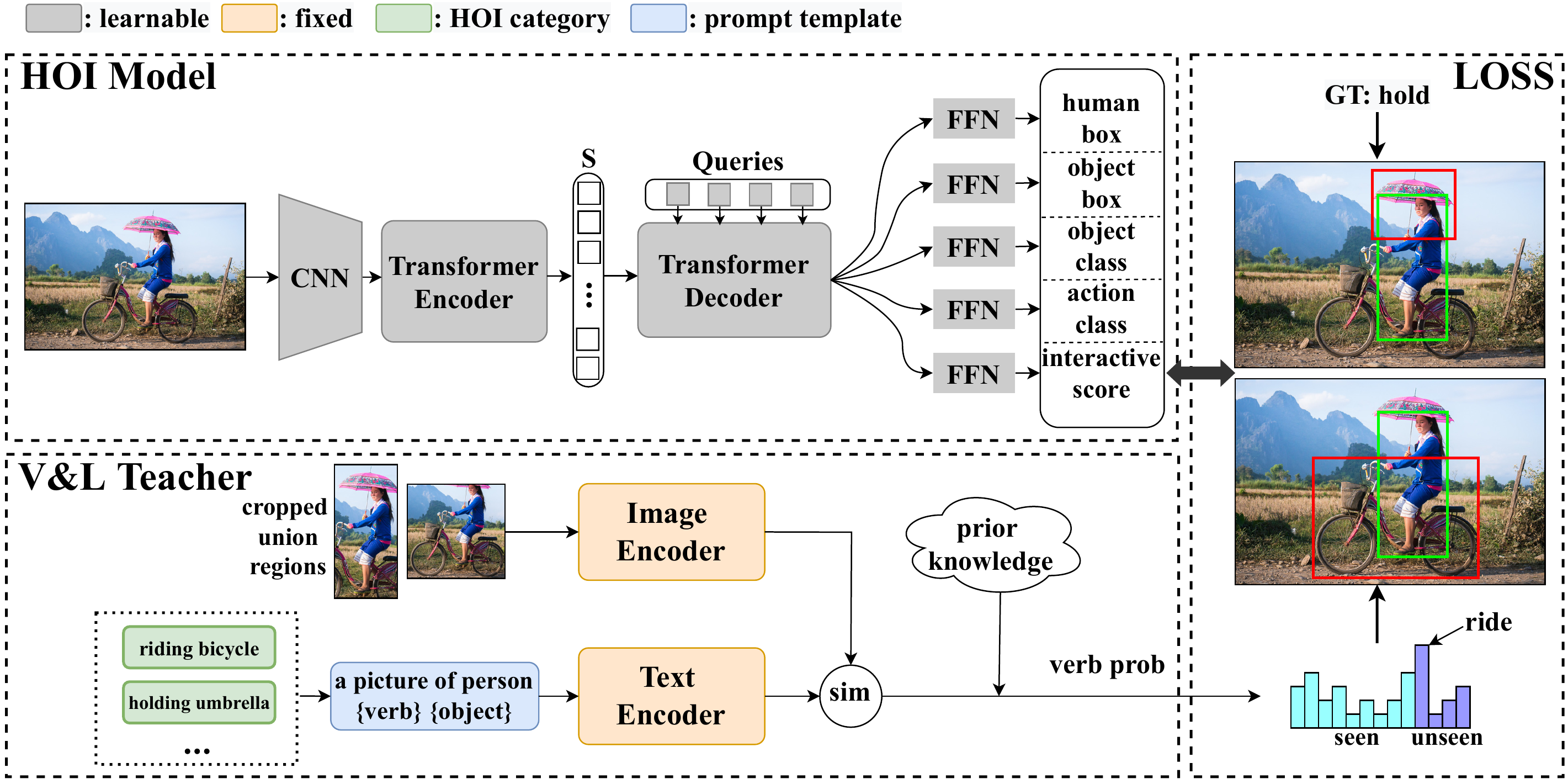} 
\caption{Overview of our EoID. 
After getting $N$ predictions from the HOI model, two-stage bipartite matching is used to select the best matched predictions with ground truth human-object pairs. Then we train the model with the selected predictions to learn the distribution of action probability from the pretrained V\&L teacher as well as the seen ground truth to achieve zero-shot HOI classification.}
\vspace{-0.35cm}
\label{fig:frame}
\end{figure*}

\section{Preliminary}
\textbf{Problem Formulation}: Denote $\mathcal{A}_S=\left\{a_1, \cdots, a_k \right\}$ as a set of the seen action categories, and $\mathcal{A}_U=\left\{a_{k+1}, \cdots, a_n \right\}$ as a set of unseen actions. Let $I$ denote an input image, with corresponding labels $\mathcal{T}=\{\mathcal{B}, \mathcal{Y}\}$ where $\mathcal{B}$ is a set of bounding boxes including human boxes $b_h$ and object boxes $b_o$, and $\mathcal{Y}$ denote a set of known HOI triplets. Each $y=\langle b_{h}, b_{o}, a\rangle$ in $\mathcal{Y}$ is an HOI triplet, where $b_{h}$ and $b_{o}$ are the elements in the set $\mathcal{B}$, and $a\in \mathcal{A}_S$.

For our HOI model, the bounding boxes are trained in a paired manner. It requires to construct all possible human-object pairs one by one between human and objects in $\mathcal{B}$. The constructed pairs  present in the annotated HOI triplets set of $\mathcal{Y}$ denote \textbf{\textit{seen}} pairs (or \textbf{\textit{known}} pairs), $y_s=\langle b_h, b_o, a\rangle$ , and the others absent in $\mathcal{Y}$ denote \textbf{\textit{unknown}} pairs, $y_u=\langle b_h, b_o, \varnothing\rangle$. The unknown pairs consist of the \textbf{\textit{unseen}} pairs with potential interaction and the \textbf{\textit{non-interactive}} pairs.
Finally, our goal is to detect all interactive \textbf{\textit{seen}} and \textbf{\textit{unseen}} pairs, and also recognize their actions.

\noindent \textbf{Transformer-based HOI models}: Most of existing SOTA HOI models~\cite{zou2021end, tamura2021qpic, zhang2021mining} are end-to-end transformer-based models. First, the input image $I$ and learnable query vectors $Q_e$ are fed to an HOI model to predict human-object bounding boxes pairs and the corresponding actions. The paradigm is formulated as,
$\hat{y} = Transformer(I, Q_e)$,
where $\hat{y}$ is the prediction. During training, a bipartite matching algorithm is adopted to match predictions with the best ground truth by the Hungarian algorithm, as follows,
\begin{equation}
\centering
\hat{\sigma}=\underset{\sigma \in \Theta_{N}}{\arg \min } \sum_{i=1}^{N} \mathcal{H}_{\text {match }}\left(y_{i}, \hat{y}_{\sigma(i)}\right)\label{qe:match}, 
\end{equation}
where $y_i\in \mathcal{\widetilde{Y}}$,  $\mathcal{\widetilde{Y}} = \{y_1, \cdots, y_M, \varnothing_{M+1}, \cdots, \varnothing_N\}$ denotes the $M$ ground truth pairs padded with $N-M$ no-pairs $\varnothing$, $\{\hat{y}_i\}^N_{i=1}$ denotes the set of $N$ predictions, and $\Theta_{N}$ is a search space for a permutation of $N$ elements. $\mathcal{H}_{match}$ is the matching cost~\cite{carion2020end}  between ground truth $y_i$ and a prediction with index $\sigma(i)$, which consists of four types of costs: the box-regression cost $\mathcal{H}_b$, intersection-over-union (IoU) cost $\mathcal{H}_u$, object-class cost $\mathcal{H}_c$, and action-class cost $\mathcal{H}_a$, as follows,
\begin{equation}
\centering
\mathcal{H}_{\text {match }} = \mathcal{H}_b+\mathcal{H}_u+\mathcal{H}_c+\mathcal{H}_a.
\end{equation}
 Finally, the losses of the matched pairs are optimized by a \textit{Hungarian loss}, which can be formulated as:
\begin{equation}
\begin{split}
\label{eq:loss}
\centering
\mathcal{L}_H=\sum_{i=1}^{N}\sum_{\mathcal{L}\in \Omega}\{\mathds{1}_{\{y_i\neq \varnothing\}}\mathcal{L}(\hat{y}_{\hat{\sigma}(i)}, y_i)+\mathds{1}_{\{y_i= \varnothing\}}\mathcal{L}(\hat{y}_{\hat{\sigma}(i)}, \varnothing)\},
\end{split}
\end{equation}
where $\Omega=\{\mathcal{L}_{b}, \mathcal{L}_{u}, \mathcal{L}_{c}, \mathcal{L}_{a}\}$, $\mathcal{L}_{b}=\mathcal{L}^{(h)}_{b}+\mathcal{L}^{(o)}_{b}$ is the box regression loss, $\mathcal{L}_{u}=\mathcal{L}^{(h)}_{u}+\mathcal{L}^{(o)}_{u}$ is the intersection-over-union loss, $\mathcal{L}_{c}$ is the object-class loss and $\mathcal{L}_{a}$ the action-class loss, following QPIC~\cite{tamura2021qpic} and CDN~\cite{zhang2021mining}. 


\section{Method}

\begin{figure*}[!t]
\centering
\includegraphics[width=0.75\textwidth]{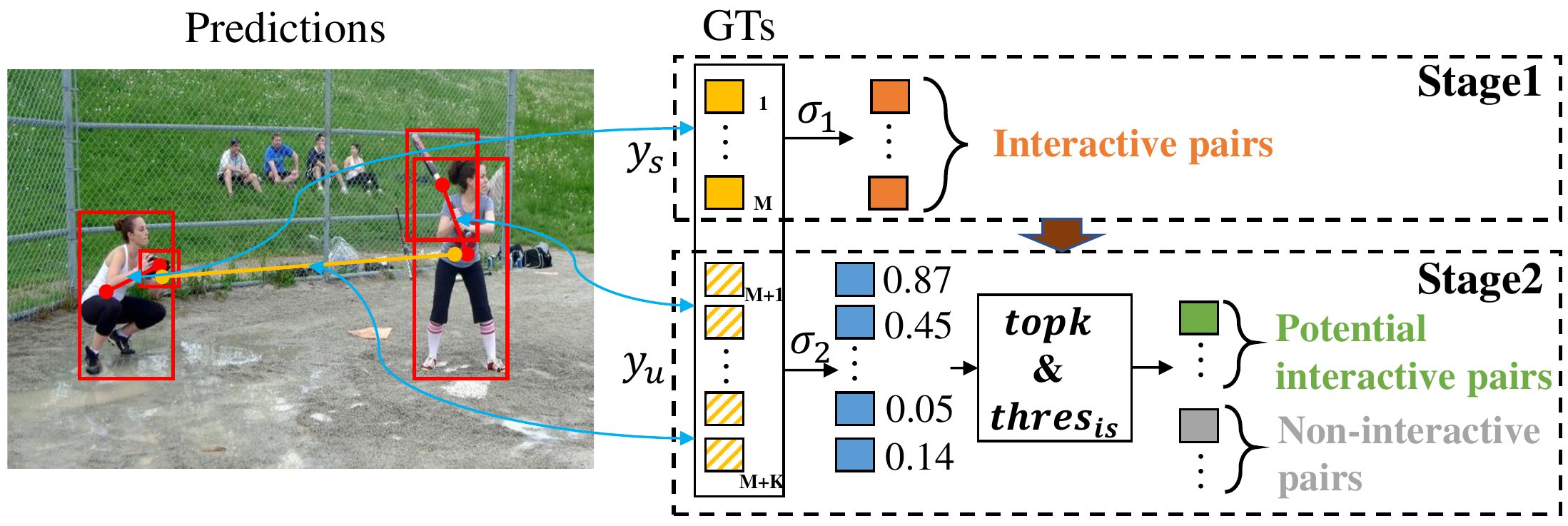} 
\caption{The two-stage bipartite matching algorithm. Given $N$ predictions, we first match all the predictions with $M$ seen pairs ($y_s$) as \textbf{interactive pairs} in the first stage. In the second stage, we use the remaining $N-M$ predictions to match the $K$ unknown pairs ($y_u$). And we select $topk$ predictions that have interactive score greater than interactive threshold $thres_{is}$ from the matched $K$ predictions as \textbf{potential interactive pairs}, and the others as \textbf{non-interactive pairs}. The non-interactive pairs and the remaining $N-M-K$ unmatched predictions will be regarded as \textbf{no-pairs}.}
\vspace{-0.35cm}
\label{fig:2stage}
\end{figure*}

\subsection{Overall Architecture}\label{m1}
Fig.~\ref{fig:frame} illustrates an overview of our EoID. We benchmark on transformer-based model CDN~\cite{zhang2021mining}. Given an image $I$, the CDN first encodes $I$ into visual feature sequence $S$, then the cascaded human-object decoder and interaction decoder are assigned to decode the visual feature sequence into $N$ predictions, including human-object bounding boxes pairs ($b_h$, $b_o$) and action vectors $\boldsymbol{a}$ with corresponding interactive scores $s_{is}$ from a set of queries $Q_e$. 
For the vision and language teacher, the union regions of human-object pairs are cropped to extract the visual features, and prompt engineering is adopted before calculating the semantic features of the HOI texts. The cosine similarity between them combined with the prior knowledge represents the action probability for these predictions.
Then a two-stage bipartite matching algorithm is applied to select the predictions that best match the ground truth human-object pairs for the seen pairs and the unknown pairs. These selected predictions from the two-stage bipartite matching algorithm are used to train bounding boxes regressive branches, object classification branch, interactive score branch, and action classification branch. Finally, we train the model to learn the distribution of action probability from the pretrained vision and language teacher as well as the seen ground truth to achieve zero-shot HOI classification.

\subsection{Learning to Detect Potential Interactive Pairs}
\label{m2}

%
%
 Exhaustively traversing all possible human-object pairs is computationally infeasible and might introduce excessive training noise.
%
As a result, the first challenge in our work is to detect potential action-agnostic interactive pairs in both training and inference. We address this issue by introducing an interactive score module and a two-stage bipartite matching algorithm.

\vspace{0.2cm} 
\noindent\textbf{Interactive Score Module}:
To distinguish interactive and non-interactive pairs in the predictions, it is a natural idea to adopt the naive implementation of the interactive score module in CDN. However, it only considers the predictions matching seen pairs $y_s$ as interactive pairs and the others as non-interactive pairs, which suppresses the detection of unseen pairs.

Unlike CDN, we apply the interactive score head from the interaction decoder instead of the human-object decoder to detect unseen pairs. Specifically, given $N$ predictions, we first apply a two-stage bipartite matching algorithm to select $M$+$topk$ predictions which best match the $M$ seen pairs and $topk$ unknown pairs. 
Then we train the interactive score module by rewarding the $M+topk$ matched predictions which have human and object IoUs with the matched ground truth pairs are greater than $0.5$, and penalizing the others. For the rest $N-M-topk$ predictions with any unknown pairs that satisfy the above condition, their loss is omitted during the optimization.

\vspace{0.2cm} 
\noindent\textbf{Two-stage Bipartite Matching Algorithm}: Similar to OW-DETR~\cite{gupta2021ow}, we apply a two-stage bipartite matching algorithm to match the seen pairs and the unknown pairs respectively, as shown in Fig.\,\ref{fig:2stage}. We first match $M$ predictions with the seen pairs as interactive pairs based on the box-regression cost $H_b$, IoU cost $H_u$, object-class cost $H_c$, and action-class cost $H_a$, as follows,
\begin{equation}
\centering
\hat{\sigma}_1=\underset{\sigma_1 \in \Theta_{N}}{\arg \min } \sum_{i=1}^{N} \mathcal{H}^1_{\text {match}}\left(y_{i}, \hat{y}_{\sigma_1(i)}\right)\label{qe:match1}, 
\end{equation}
where $y_i\in \mathcal{\widetilde{Y}}_1 = \{y_1, \cdots, y_M, \varnothing_{M+1}, \cdots, \varnothing_N\}$, and $\mathcal{H}^1_{\text {match}} = \mathcal{H}_b+\mathcal{H}_u+\mathcal{H}_c+\mathcal{H}_a$. Then the $N-M$ predictions not selected by the first stage matching will be used for the second stage matching. Since ground truth actions are not available for the unknown pairs, we implement the second stage matching only based on the box-regression cost $H_b$, IoU cost $H_u$, object-class cost $H_c$. The second stage matching process can be formulated as follows,
\begin{equation}
\begin{aligned}
\hat{\sigma}_2=\underset{\sigma_2 \in \Theta_{N-M}}{\arg \min } \sum_{i=M+1}^{N} \mathcal{H}^2_{\text {match}}\left(y_{i}, \hat{y}_{\sigma_2(i)}\right)\label{qe:match2}, \\
\end{aligned}
\end{equation}
where $\mathcal{H}^2_{\text {match}} = \mathcal{H}_b+\mathcal{H}_u+\mathcal{H}_c$, $y_i\in\mathcal{\widetilde{Y}}_2 = \{y_{M+1}, \cdots, y_{M+K}, \varnothing_{M+K+1}, \cdots, \varnothing_N\}$, and $K$ is the number of unknown pairs. We select the $topk$ predictions which have interactive scores $s_{is}$ greater than interactive threshold $thres_{is}$ as potential interactive pairs, the others as non-interactive pair. So we can get the $M+topk$ predictions with $\hat{\sigma} = \hat{\sigma}_1 \cup topk(\hat{\sigma}_2)$. The non-interactive pairs and the remaining $N-M-K$ unmatched predictions will be regarded as no-pairs. The no-pairs and $M+topk$ predictions combined with matched ground truth will be used to train bounding boxes regressive branch, object classification branch, interactive score branch and action classification branch.

Such a strategy will help the model learn from the seen pairs to discriminate whether there exists interaction between human-object pairs at the early training stage, and gradually introduce potential interactive pairs for learning. Different from GEN-VLKT~\cite{liao2022gen} that simply detects known pairs, our method also additionally detects potential interactive pairs which not exist in training set and can also be applied to scale up the action sets on object detection dataset.

\subsection{Knowledge Distillation from CLIP}
\label{m3}

After detecting potential interactive human-object pairs, we need to identify the corresponding action happening between the human and object. For this purpose, we transfer the interaction knowledge from the pretrained V\&L model CLIP (teacher) into the HOI model (student)  
via knowledge distillation similar to ViLD~\cite{gu2021open} and GEN-VLKT~\cite{liao2022gen}. In contrast to global image-level distillation in GEN-VLKT, we adopt local region-level distillation to deal with multiple human-object pairs in one image. In order to avoid the misalignment between the local feature of the teacher and global feature of the student, we apply logits distillation instead of feature distillation 
adopted by ViLD.

We first convert the HOI category texts, \emph{e.g.} \textit{riding bicycle}, into the prompts by feeding them into prompt template \textit{a picture of person $\left \{verb\right\}$ $\left\{object\right\}$}. 
Then we encode these prompts to generate the text embeddings $t_e$ offline by the CLIP text encoder $T$. For the $M+topk$ matched pairs of $I$, we crop the human-object union regions, and feed the preprocessed ones into the CLIP image encoder $V$ to generate the image embeddings $v_e$.
Then, we compute cosine similarities between the image and text embeddings, as $s_i = v_e^Tt_e^i \mathbin{/} (\|v_e\|\cdot \|t_e^i\|)$. According to the prior knowledge~\cite{chao2018learning}, we select the valid actions which is able to interact with the object for each union region. This makes the model pay more attention to the learning of the current human-object pair, and avoid the interference of other human-object interactions in the union-box. We apply a softmax activation on similarities of these HOI categories to get the probability distribution $\mathbf{p}$ of the actions in $\mathcal{A}_S + \mathcal{A}_U$ for each of union regions. The process can be formulated as follows, 
\begin{equation}
\begin{aligned}
p_i &= \frac{e^{\gamma\,s_i m_i}}{\sum_{j=1}^{n}e^{\gamma\,s_j m_j}}, \\
\mathrm{s.t.}~m_i&= \begin{cases}1, & \text { if } \textit { valid } \\ -\infty, & \text { if } \textit { invalid }\end{cases}, \\
\end{aligned}
\end{equation}
where $p_i$ is the probability of the action, $\gamma$ is a scalar hyper-parameter and $m_i$ is a correct coefficient to eliminate invalid HOI categories~\cite{chao2018learning}.
Finally, we train the model to fit this probability distribution $\mathbf{p}$ of the actions in $\mathcal{A}_S + \mathcal{A}_U$ as well as the ground truth actions in $\mathcal{A}_S$.

\subsection{Training and Inference}
\label{m4}
\textbf{Training}: We calculate the loss with extra interactive score loss and  CLIP distillation loss, as follows,
\begin{equation}
\mathcal{L}_{total} = \mathcal{L}_H + \lambda_{is} \mathcal{L}_{is} + \lambda_{clip} \mathcal{L}_{clip},\\
\end{equation}
where $\mathcal{L}_H$ is computed by Eq.\,\ref{eq:loss}, $\mathcal{L}_{is}$ is the interactive score loss, and $\mathcal{L}_{clip}$ is the CLIP distillation loss. The $\mathcal{L}_{is}$ term adopts cross entropy loss, and the $\mathcal{L}_{clip}$ term adopts binary cross entropy loss. $\lambda_{is}$ and $\lambda_{clip}$ are the hyper-parameters. 

\vspace{0.2cm} 
\noindent
\textbf{Inference}: After distilling the action knowledge from CLIP, we only keep the learned CDN model for inference, avoiding extra computation cost. The post-process of our method remains unchanged as CDN.

\section{Experiments}
\subsection{Experimental Setup}\label{sec:set}
\textbf{Datasets and Evaluation Metrics}:
We perform our experiments on two HOI detection benchmarks: HICO-DET~\cite{chao2018learning} and V-COCO~\cite{gupta2015visual}. 
We follow the standard evaluation~\cite{chao2018learning} to use the mean average precision ($m$AP) as the evaluation metric. A HOI triplet is considered as a true positive when (1) the predicted object and action categories are correct, and (2) both the predicted human and object bounding boxes have intersection-over-union (IoU) with a ground truth greater than 0.5.

\setlength{\tabcolsep}{4pt}
\begin{table}[!t]
\begin{center}
\caption{The recall on unseen pairs (U-R). We compare the models trained with different interactive score modules and supervision sources. Our method achieves comparable recall for unseen pairs on HICO-Det test set. Results show that our method can efficiently detect potential interactive pairs.}
\label{table:ar}
\begin{tabular}{lcccc}
\toprule
Method & Supervision & U-R@3 & U-R@5 & U-R@10 \\
\midrule
 CDN & seen & 61.47 & 66.68 & 71.70 \\
 EoID & seen & 64.72 & \textbf{71.45} & 76.79 \\
 EoID & seen+$topk$ & \textbf{65.25} & 71.16 & \textbf{77.20} \\
\midrule
 CDN & full & 67.03 & 73.41 & 78.64 \\
\bottomrule
\end{tabular}
\end{center}
\end{table}
\setlength{\tabcolsep}{1.4pt}

\noindent
\textbf{Zero-shot Setups}:
We conduct experiments on HICO-Det: unseen combination scenario (UC)~\cite{bansal2020detecting}, Rare-first UC (RF-UC)~\cite{hou2020visual}, Non-rare-first UC (NF-UC)~\cite{hou2020visual}, unseen action scenario (UA)~\cite{liu2020consnet} and UV~\cite{liao2022gen}. 
Details are shown in the supplementary material.

\noindent
\textbf{Implementations}:
We benchmark on the CDN~\cite{zhang2021mining} and use the same settings for all models unless explicitly specified. The query number $N$ is 64. The loss weights $\lambda_{bbox}, \lambda_{giou}$, $\lambda_{c}$, $\lambda_{is}$, $\lambda_{a}$ and $\lambda_{clip}$ are set to 2.5, 1, 1, 1, 1.6 and 700 respectively. 
For simplicity, the decoupling dynamic re-weighting in CDN is not used.
For CLIP, we use the public pretrained model\,\footnote{\url{https://github.com/openai/CLIP}.}, with an input size of $224\times224$, and $\gamma$=100. The cropped union regions are preprocessed by square padding and resizing. We feed prompt engineered texts to the text encoder of CLIP with a prompt template \emph{a picture of person $\left\{verb\right\}$ $\left\{object\right\}$}. Experiments are conducted on 4 Tesla V100 GPUs, with a batch size of 16.

\subsection{Learning to Detect Unseen Pairs}\label{sec:learn}
We first show our framework can detect potential interactive pairs. We compare the models trained with different interactive score (IS) modules and supervision sources. Table\,\ref{table:ar} shows the top-k recall (U-R@K) of unseen pairs on HICO-Det test set. Even though training only with seen pairs (seen), the recall on unseen pairs of our model outperforms the CDN with naive IS module by a large margin. After introducing the potential pairs ($topk$) for training, the recall is further improved and shows comparable results with the fully-supervised model. The result shows that our method can detect more potential interactive pairs than the CDN.

\begin{figure}[t]
\centering
\includegraphics[width=0.98\columnwidth]{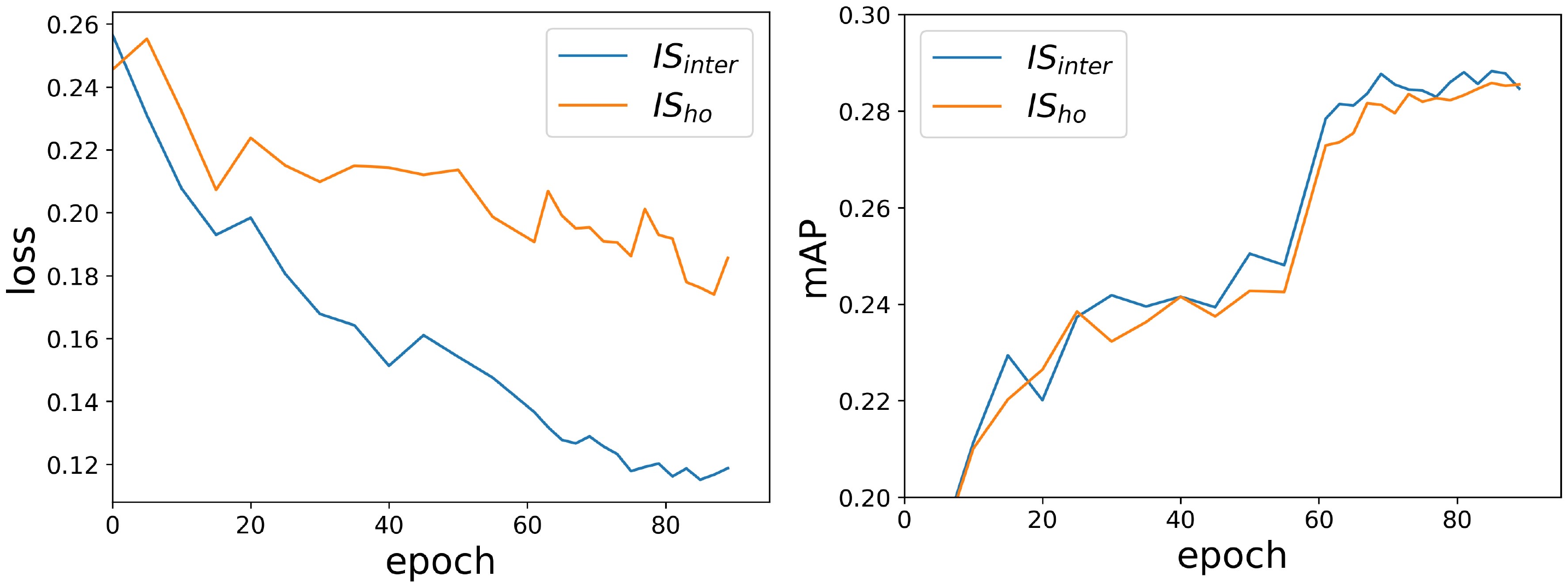} 
\caption{The curves of the interactive score loss (left) and the $m$AP (right), compared between the models with IS branch from interactive decoder and human-object decoder. The model with IS branch from interactive decoder shows faster convergence and better performance.}
\label{fig:is}
\end{figure}

We also compare the curves of the interactive score loss $\mathcal{L}_{is}$ and the overall $m$AP with IS branch from the interactive decoder ($IS_{inter}$) and human-object decoder ($IS_{ho}$), as shown in Fig.\,\ref{fig:is}. Compared to the model with $IS_{ho}$, the interactive score loss of the model with $IS_{inter}$ converges faster, and also shows a better overall performance. These results demonstrate the interactive decoder is more capable of extracting interactive information.

\setlength{\tabcolsep}{4pt}
\begin{table}[t]
\begin{center}
\caption{We replace the action classifier of the converged CDN  with CLIP to validate the zero-shot transferability of CLIP on HOI classification, with or without prior knowledge. Using CLIP for HOI classification shows the competitive performance with the full-supervised model on rare HOI categories of HICO-Det. } 
\label{table:clip}
\begin{tabular}{lcccc}
\toprule
Method & Full & Rare & Non-rare \\
\midrule
 CLIP & 21.11 & 26.02 & 19.64 \\
 CLIP w/ prior & 21.45 & 26.42 & 19.97 \\
\midrule
Full-Supervised(seen + unseen) & 31.11 & 26.49 & 32.49 \\
\bottomrule
\end{tabular}
\end{center}
\end{table}
\setlength{\tabcolsep}{1.4pt}

\subsection{Zero-Shot Transferability of CLIP on HOI Classification}\label{sec:clip}
We replace the action classifier of the converged CDN model with CLIP to validate the zero-shot transferability of CLIP on HOI classification. The prior knowledge is also applied to avoid the dispersion of probability distribution of CLIP to some invalid actions. The CLIP model is RN50x16 and the CDN model is CDN-S. We report the $m$AP on the full set of HOI classes (Full), a rare set of the HOI classes (Rare) that have less than 10 training instances, and a non-rare set of the other HOI classes (Non-rare). As shown in Table\,\ref{table:clip}, the HOI model with a fixed CLIP classifier only has a small degradation with the fully supervised model on \textit{Rare} categories. This gap is further reduced when prior knowledge is adopted, which shows the competitive performance with the full-supervised model. However, the performances of \textit{Full} and \textit{Non-rare} are not effective enough. Note that the CLIP classifies all the predicted pairs without considering whether it is interactive or not. There is still much improvement space for the overall performance and the inference speed.

\subsection{Ablation Studies}
\label{sec:abl}
\label{abl}
We perform ablation experiments in Sec.\,\ref{abl}. Unless otherwise specified, the CLIP model used here is RN50x16, the CDN model is CDN-S, the $topk$ and $thres_{is}$ are set to 3 and 0.5 respectively.
All ablation results are evaluated on the HICO-Det test set.

\noindent
\textbf{Ablation of variants}:
As shown in Sec.\,\ref{sec:clip}, the degradation on \textit{Full} and \textit{Non-rare} indicates the existence of noise from CLIP, which may lead to poor performance on the \textit{Seen} category.
In addition, extra loss terms are introduced into the framework which leads to the problem of convergence difficulties. We study the methods to overcome the above challenges: 1) we only distill CLIP to the HOI actions in $\mathcal{A}_U$ under UA setting, short as \textit{$\mathcal{A}_U$ only}; 2) we use \textit{detach technique}
to cut off the back-propagation of gradients between the human-object decoder and interaction decoder. As shown in Table\,\ref{table:trick}, we can combine the \textit{$\mathcal{A}_U$ only} to alleviate the impact of the noise from CLIP, which improves the \textit{Seen} and \textit{Unseen} by $1.05\%$ and $0.14\%$ $m$AP respectively. With the \textit{detach technique} adopted, the best \textit{Full} performance is obtained to $29.22\%$ $m$AP.

\setlength{\tabcolsep}{4pt}
\begin{table}[t]
\begin{center}
\caption{Ablation of variants. We study the methods to cope with the problems occurred in Sec.\,\ref{sec:clip}.}
\label{table:trick}
\begin{tabular}{cccccc}
\toprule
$A_U$ only & detach & Full & Seen & Unseen \\
\midrule
  & & 27.93 & 29.15 & 21.84  \\
 \Checkmark & & 28.83 & 30.20 & 21.98  \\
  & \Checkmark & 28.68 & 29.88 & 22.71  \\
 \Checkmark & \Checkmark & \textbf{29.22} & \textbf{30.46} & \textbf{23.04} \\
\bottomrule
\end{tabular}
\end{center}
\vspace{-0.35cm}
\end{table}
\setlength{\tabcolsep}{1.4pt}

\setlength{\tabcolsep}{1.4pt}
\begin{table}[t]
\begin{center}
\caption{Ablation studies. We perform ablation experiments to study the impact of $topk$ and $thres_{is}$.}
\label{table:ablation}
\begin{subtable}[t]{0.495\linewidth}
\centering
\caption{Impact of $topk$.}
\label{abl:topk}
\begin{tabular}{lccc}
\toprule
$topk$ & Full & Seen & Unseen  \\
\midrule
 1 & 28.45 & 29.75 & 21.96  \\
 3 & \textbf{28.83} & \textbf{30.20} & 21.98   \\
 5 & 28.56 & 29.85 & 22.14   \\
 10 & 28.45 & 29.69 & \textbf{22.29}  \\
\bottomrule
\end{tabular}
\end{subtable}
\begin{subtable}[t]{0.495\linewidth}
\caption{Impact of $thres_{is}$.}
\label{abl:thres}
\begin{tabular}{lccc}
\toprule
$thres_{is}$ & Full & Seen & Unseen \\
\midrule
 0.1 & 28.50 & 29.80 & 22.01 \\
 0.3 & 28.48 & 29.59 & \textbf{22.93} \\
 0.5 & \textbf{28.83} & \textbf{30.20} & 21.98 \\
 0.7 & 28.08 & 29.11 & 22.90 \\
 0.9 & 28.44 & 29.99 & 20.75 \\
\bottomrule
\end{tabular}
\end{subtable} \\
\end{center}
\vspace{-0.5cm}
\end{table}
\setlength{\tabcolsep}{1.4pt}

\noindent
\textbf{Impact of $topk$}:
We compare models with different $topk\in\{1,3,5,10\}$, and results are shown in Table\,\ref{abl:topk}. The model with $topk$=3 obtains the best performance on \textit{Full} and \textit{Seen}. With the growth of the $topk$, the model obtains a better performance on \textit{Unseen} while the overall $m$AP starts to drop for too much noise.

\setlength{\tabcolsep}{3pt}
\begin{table}[t]
\footnotesize
\begin{center}
\caption{Zero-shot HOI Detection results on HICO-DET dataset. UC and UA(UV) denote unseen action-object combination and unseen action scenarios respectively, RF-UC and NF-UC denote rare-first and non-rare-first UC scenarios. Our method outperforms all the other methods by a large margin. The $*$ indicates that the model training without using \textit{$A_U$ only} and \textit{detach technique}. The \ddag\space denotes our implementation.}
\label{table:sota}
\begin{tabular}{lcccc}
\toprule
Method & Type & Full & Seen & Unseen\\
\midrule
VCL & RF-UC & 21.43 & 24.28 & 10.06 \\
ATL & RF-UC & 21.57 & 24.67 & 9.18 \\
FCL & RF-UC & 22.01 & 24.23 & 13.16 \\
GEN-VLKT & RF-UC & \textbf{30.56} & \textbf{32.91} & 21.36 \\
\rowcolor{lightgray} baseline & RF-UC & 28.46 & 30.80 & 19.10 \\
\rowcolor{lightgray} EoID & RF-UC & 29.52 & 31.39 & \textbf{22.04} \\
\midrule
VCL & NF-UC & 18.06 & 18.52 & 16.22 \\
ATL & NF-UC & 18.67 & 18.78 & 18.25 \\
FCL & NF-UC & 19.37 & 19.55 & 18.66 \\
GEN-VLKT & NF-UC & 23.71 & 23.38 & 25.05 \\
\rowcolor{lightgray} baseline & NF-UC & 23.93 & 25.18 & 18.94 \\
\rowcolor{lightgray} EoID & NF-UC & \textbf{26.69} & \textbf{26.66} & \textbf{26.77} \\
\midrule
Functional & UC & 12.45$\pm$0.16 & 12.74$\pm$0.34 & 11.31$\pm$1.03 \\
ConsNet & UC & 19.81$\pm$0.32 & 20.51$\pm$0.62 & 16.99$\pm$1.67 \\
GEN-VLKT$^{\ddag}$ & UC & 25.23$\pm$0.59 & 27.16$\pm$0.88 & 20.64$\pm$0.89 \\
\rowcolor{lightgray} baseline & UC & 26.57$\pm$0.43 & 28.65$\pm$0.58 & 18.24$\pm$1.02 \\
\rowcolor{lightgray} EoID & UC & \textbf{28.91$\pm$0.27} & \textbf{30.39$\pm$0.40} & \textbf{23.01$\pm$1.54} \\
\midrule
ConsNet & UA & 19.04 & 20.02 & 14.12 \\
GEN-VLKT$^{\ddag}$ & UA & 26.28 & 28.72 & 20.85 \\
\rowcolor{lightgray} baseline & UA & 26.53 & 28.77 & 15.30 \\
\rowcolor{lightgray} EoID* & UA & 27.93 & 29.15 & 21.84 \\
\rowcolor{lightgray} EoID & UA & \textbf{29.22} & \textbf{30.46} & \textbf{23.04} \\
\midrule
GEN-VLKT & UV & 28.74 & 30.23 & 20.96 \\
\rowcolor{lightgray} EoID & UV & \textbf{29.61} & \textbf{30.73} & \textbf{22.71} \\
\bottomrule
\end{tabular}
\end{center}
\vspace{-0.5cm}
\end{table}
\setlength{\tabcolsep}{1.4pt}

\noindent
\textbf{Impact of $thres_{is}$}:
We compare models with different $thres_{is}\in\{0.1,0.3,0.5$, $0.7,0.9\}$, and results are shown in Table\,\ref{abl:thres}. The models with a larger $thres_{is}$ will eliminate more non-interactive pairs, which may results in a lager precision but a lower recall of unseen pairs and vice versa. The best performance of \textit{Full} and \textit{Seen} are obtained when adopting a proper $thres_{is}$=0.5.

\subsection{Zero-Shot HOI Detection}
We compare our method with state-of-the-art models on the HICO-Det test set under RF-UC, NF-UC, UC, UA and UV settings in Table\,\ref{table:sota}. The compared models include: 
Functional~\cite{bansal2020detecting},
VCL~\cite{hou2020visual}, ATL~\cite{hou2021atl},
FCL~\cite{hou2021detecting},
GEN-VLKT~\cite{liao2022gen}, and ConsNet~\cite{liu2020consnet}. 
Besides, we introduce a new variant of CDN by applying the consistency graph of ConsNet (CDN+ConsNet, details in supplementary material) as our \textbf{baseline} to validate whether the better performance is obtained by the better backbone. 
As shown in Table\,\ref{table:sota}, our method outperforms the previous SOTA on \textit{Unseen} under five various zero-shot settings, which valid the effectiveness of the proposed EoIDs. Compared to GEN-VLKT, the slightly lower performance on \textit{Seen} at the RF-UC setting may be due to the different HOI models, which is evaluated in GEN-VLKT~(Tab. 4(a)).
Even with a weaker HOI model CDN, our methods still outperform GEN-VLKT at the remaining settings.
Note that, our best \textit{Unseen} performance is competitive with the fully-supervised method~\cite{zou2021end} ($23.04\%$ v.s. $23.46\%$) and these experiments indicate the effectiveness of our proposed zero-shot HOI detection framework.

\setlength{\tabcolsep}{3pt}
\begin{table}[t]
\begin{center}
\caption{Transfer to object detection datasets. We study the performance of our method on V-COCO with the bounding box annotations from COCO. Our method can transfer to datasets with only bounding boxes annotated to further scale up existing HOI categories.}
\label{table:hvco}
\begin{tabular}{lcccc}
\toprule
Training source & Method & Full & Seen & Unseen \\
\midrule
 HICO only & CDN & - & 35.15 & - \\
 HICO+pseudo-V-COCO & CDN & - & 38.13 & - \\
 HICO+pseudo-V-COCO & EoID & 40.39 & 38.13 & 47.15 \\
 \midrule
V-COCO(full) & CDN & 56.43 & 54.56 & 62.05 \\
\bottomrule
\end{tabular}
\end{center}
\vspace{-0.5cm}
\end{table}
\setlength{\tabcolsep}{1.4pt}

\subsection{Scaling up Action Sets via Object Detection Datasets}\label{sec:V-COCO}
Current HOI detectors are limited in action size due to the small scale of HOI detection datasets. Meanwhile, large-scale object detection datasets contain potential unknown interaction pairs. As a result, we introduce a more difficult but practical experiment, in which object detection datasets are merged to further scale up the existing HOI categories. For this purpose, we study the performance of our method on V-COCO with the bounding box annotations from COCO. Specially, we compare the models trained on HICO-Det dataset (HICO only), HICO-Det with pseudo-V-COCO and full V-COCO dataset, where the pseudo-V-COCO consists of bounding boxes from COCO, pseudo-seen action labels from the predictions of the CDN trained on HICO-Det dataset. We test on V-COCO test set, and set the overlap actions between HICO-Det and V-COCO as \textit{Seen} and the others as \textit{Unseen}. More details are elaborated in Appendix. Results are shown in Table\,\ref{table:hvco}, and our method has an about $16\%$ overall $m$AP gap compared to the fully-supervised method, while a smaller gap on \textit{Unseen} categories. This experiment shows that our method can transfer to bounding boxes annotated datasets to further scale up the existing HOI categories.




\section{Conclusions}

In this work, we present EoID, an end-to-end zero-shot HOI detection framework via knowledge distillation from multimodal vision-language embeddings.
Our method first detects potential action-agnostic interactive human-object pairs by applying a two-stage bipartite matching algorithm and an interactive score module.
Then a zero-shot action classification is applied to identify novel HOIs.
The experiments demonstrate that our detector is able to detect unseen pairs, which benefits the recognition of unseen HOIs.
%
Our method outperforms the previous SOTAs under four zero-shot settings and shows a promising generalization to utilize large-scale detection datasets to scale up the action sets.

\section*{Acknowledgement}
This work was supported by the National Science Fund for Distinguished Young Scholars (No.62025603), the National Natural Science Foundation of China (No. U21B2037, No. 62176222, No. 62176223, No. 62176226, No. 62072386, No. 62072387, No. 62072389, and No. 62002305), Guangdong Basic and Applied Basic Research Foundation (No.2019B1515120049), and the Natural Science Foundation of Fujian Province of China (No.2021J01002).
\bibliography{aaai23}

\end{document}